\pdfoutput=1

\documentclass[11pt]{article}
\usepackage[]{emnlp2021}

\usepackage{times}
\usepackage{latexsym}

\usepackage[T1]{fontenc}
\usepackage[utf8]{inputenc}

\usepackage{microtype}

\usepackage{times,latexsym}
\usepackage{url}
\usepackage{paralist}

\usepackage{xspace,mfirstuc,tabulary}

\usepackage{mathptmx}
\usepackage{latexsym}

\usepackage{tabularx}
\usepackage{booktabs}
\usepackage{amsmath}
\usepackage{amssymb}
\usepackage{enumitem}
\usepackage{multirow}
\usepackage{listings}
\usepackage{newfloat}
\usepackage[framemethod=TikZ]{mdframed}
\usepackage{hyperref}
\usepackage{microtype}
\usepackage{caption}
\usepackage{arydshln}
\usepackage{pifont}
\newcommand{\cmark}{\ding{51}}%
\newcommand{\xmark}{\ding{55}}%

\DeclareFloatingEnvironment[placement={!ht},name=Example]{example}
\newcommand{\exampleref}[1]{Example~\ref{#1}}
\newcommand{\multiwoz}[0]{MultiWOZ 2.0\xspace}
\newcommand{\multiwozn}[0]{MultiWOZ 2.1\xspace}
\newcommand{\taskmaster}[0]{Taskmaster-1\xspace}
\newcommand{\schema}[0]{Schema-Guided Dialogue\xspace}
\newcommand{\augpt}[0]{AuGPT\xspace}
\newcommand{\Augpt}[0]{AuGPT\xspace}

\newcommand{\code}[1]{\texttt{#1}}

\mdfdefinestyle{ExampleFrame}{%
    nobreak=true,
    innertopmargin=4pt,
    innerbottommargin=4pt,
    innerrightmargin=4pt,
    innerleftmargin=4pt,
    outermargin=0,
    }

\title{{AuGPT}: {A}uxiliary Tasks and Data Augmentation for End-To-End\\ Dialogue with Pre-Trained Language Models}

\author{Jonáš Kulhánek,\textsuperscript{\rm 1,2,3} Vojtěch Hudeček,\textsuperscript{\rm 1} Tomáš Nekvinda\textsuperscript{\rm 1} \and Ondřej Dušek\textsuperscript{\rm 1} \\
\textsuperscript{\rm 1}Charles University, Faculty of Mathematics and Physics, Prague, Czechia \\
\textsuperscript{\rm 2}Czech Technical University in Prague, Czech Institute of Informatics, Robotics and Cybernetics\\
\textsuperscript{\rm 3}Czech Technical University in Prague, Faculty of Electrical Engineering\\
\texttt{jonas.kulhanek@cvut.cz, \{hudecek,nekvinda,odusek\}@ufal.mff.cuni.cz}
}

\begin{document}
\maketitle
\begin{abstract}
Attention-based pre-trained language models such as GPT-2 brought considerable progress to end-to-end dialogue modelling. However, they also present considerable risks for task-oriented dialogue, such as lack of knowledge grounding or diversity. To address these issues, we introduce modified training objectives for language model finetuning, and we employ massive data augmentation via back-translation to increase the diversity of the training data. We further examine the possibilities of combining data from multiples sources to improve performance on the target dataset. We carefully evaluate our contributions with both human and automatic methods. Our model substantially outperforms the baseline on the MultiWOZ data and shows competitive performance with state of the art in both automatic and human evaluation.
\end{abstract}

\section{Introduction}
Unlike traditional task-oriented systems based on modularized pipelines \cite{young2013, gao2018}, end-to-end dialogue systems integrate nearly all functionality required to hold a dialogue into a single neural network \cite{wen2017,manning2017,lei2018}, reducing error-propagation and data annotation requirements. While these systems are not yet ready for production use, they made considerable progress in recent years, especially with the advent of pre-trained neural language models (LMs) \cite{devlin2019,radford2019,zhang2020dialogpt}.
Systems such as GPT-2 finetuned by \citet{budzianowski2019} show that with an LM pre-trained on a large number of general-domain dialogues without annotation, only small amounts of data are required to perform well in a given task-oriented domain.

On the other hand, the pre-trained LMs run enormous risks. First, solely training for response generation may result in a lack of grounding for the responses, where the LM hallucinates words without any relation to the database. This has been addressed by multi-task training and auxiliary training objectives \cite{peng2020} to an extent. Second, finetuning on small datasets may reduce response diversity and fluency due to neural networks' known propensity for catastrophic forgetting \cite{greco_psycholinguistics_2019} -- the model overfits the finetuning dataset too tightly, “forgetting” the pre-trained language modeling capabilities.

This paper presents an end-to-end model for multi-domain task-oriented response generation on the MultiWOZ data \cite{budzianowski2018},\footnote{\url{https://convlab.github.io}} where we address the above problems with pre-trained LMs. \Augpt is based on the GPT-2 LM and \citet{peng2020}'s basic approach.
Our contributions can be summarized as follows:
\begin{itemize}[itemsep=0pt,topsep=2pt,leftmargin=12pt]
    \item We introduce a new dialogue consistency classification task based on subtle changes to the dialogue state (instead of fully random resampling) used as an auxiliary training objective, and we demonstrate its performance improvements.
    \item We present a novel application of token unlikelihood loss \cite{welleck2019} in task-oriented dialogue to further improve diversity of our model's responses.
    \item We apply pre-training on additional datasets and massive data augmentation using back-translation via multiple languages \cite{sennrich2016} and demonstrate that both markedly improve task-oriented dialogue performance.
    \item We compare our model to multiple baselines on MultiWOZ in a corpus-based and simulated evaluation. 
    We also include human evaluation results from a shared task competition, as well as detailed manual error analysis.
\end{itemize}
We publish our augmented training data, source code, and pre-trained models on GitHub.\footnote{\url{https://github.com/ufal/augpt}}

\section{Related Work}
\label{sec:related}

While the first attempts to build generative end-to-end task-oriented systems mimicked the traditional dialogue system components \citep{wen2017}, the task was soon recast as a sequence prediction problem in a two-stage setup. A sequence-to-sequence (seq2seq) model first generates the belief state based on dialogue context, then generates the system response based on the context and the belief state \cite[Sequicity;][]{lei2018}.

Recently, large-scale multi-domain task-oriented datasets were proposed \cite{budzianowski2018, byrne2019, rastogi2019}.
To address multiple domains, \citet{zhang2020end2end} introduce the LABES-S2S model that -- in addition to a two-stage seq2seq approach -- models belief states as discrete latent variables.
\citet{zhang2019} present DAMD, a three-stage seq2seq architecture which explicitly decodes the system action. They optimize for multiple good actions given a single belief state. \citet{qin-etal-2020-dynamic} investigate sharing of domain knowledge and performance on unseen domains.
\citet{lubis-etal-2020-lava}'s LAVA model employs reinforcement learning over latent system actions initialized using a variational autoencoder.

The line of research closest to our work makes use of large pre-trained LMs 
based on the transformer architecture \cite{vaswani2017} such as GPT-2 \cite{radford2019} or BERT \cite{devlin2019}. 
For example, \citet{wu2020}
propose finetuning BERT \cite{devlin2019} for task-oriented dialogue,
\citet{zhang2020dialogpt} extended the GPT-2 LM to model open-domain chit-chat.

We follow research initiated by \citet{budzianowski2019}, who use GPT-2 to model multi-domain task-oriented dialogues.  
Recently, three similar modifications to their model were proposed, namely SOLOIST \cite{peng2020}, SimpleTOD \cite{hosseini2020}, and the approach by \citet{ham2020}. 
Our work extends these models and proposes a novel training approach and data augmentation strategies based on back-translation \cite{edunov2018,federmann2019multilingual}. Earlier works used a single pivot language \cite{jin-etal-2018, Einolghozati2019}, whereas our work applies 10 languages to increase variability.

\section{Method}
\label{sec:method}

\begin{figure*}[htbp]
\centering
\includegraphics[width=0.8\textwidth]{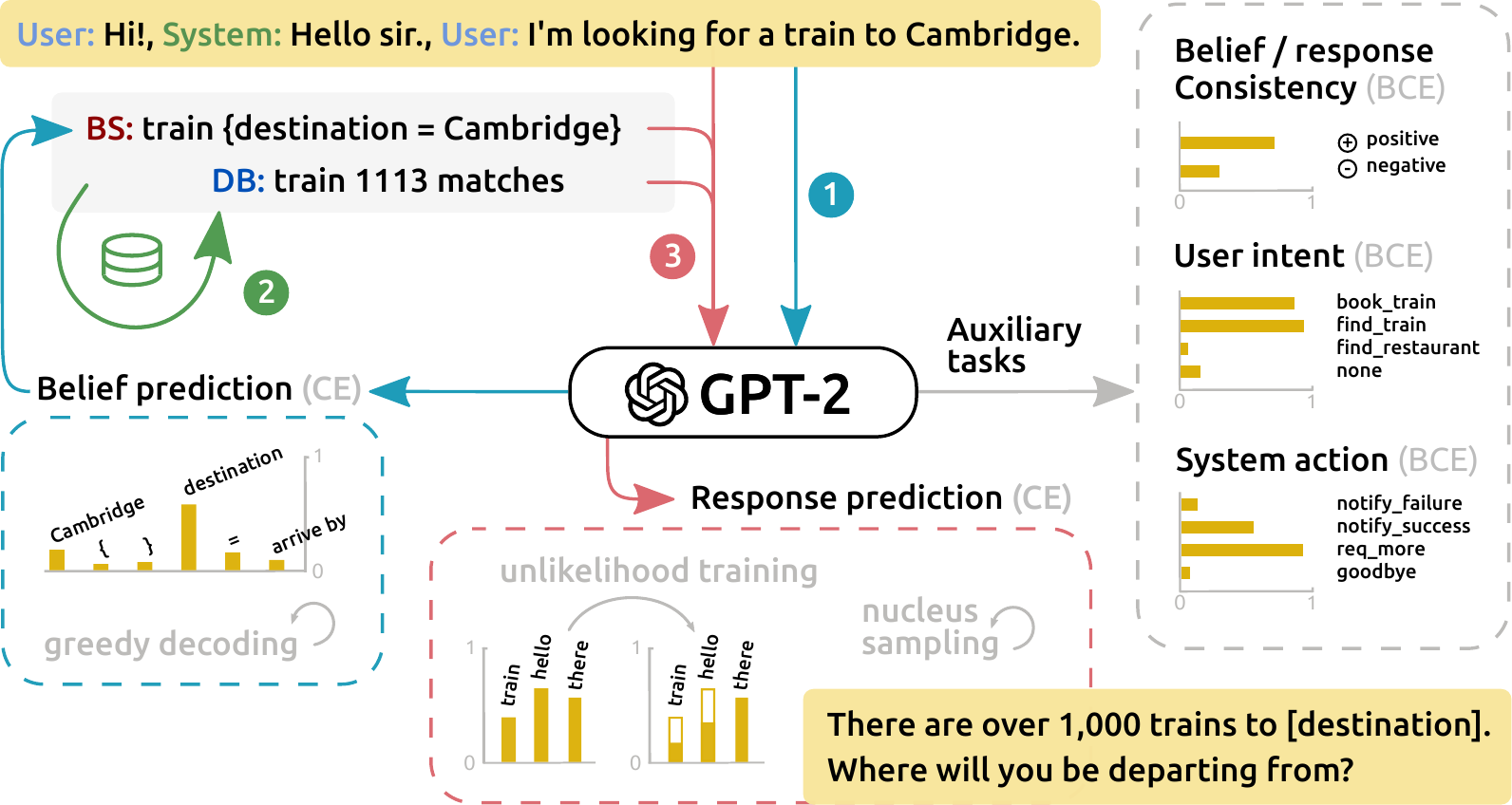}
\caption{The architecture of \augpt. The pipeline runs in two stages. First, a finetuned GPT-2 LM is used to predict a belief. Then the database results are obtained and everything is passed to the GPT-2 again to predict a final delexicalized response, along with possible auxiliary tasks (belief consistency, intent classification, system action classification). Unlikelihood loss is used for response prediction training.}
\label{fig:pipeline}
\end{figure*}

The task-oriented setting requires the dialogue system to respond adequately to the user's input and fulfill its goal, 
e.g., booking a train or requesting restaurant details. 
The system must process the user's input, keep track of the belief state (user preferences regarding individual slots, i.e., in-domain attributes) and generate a relevant response in natural language. It must also interact with a database to incorporate external information into its responses (see Figure~\ref{fig:pipeline} for an example).
Following \citet{budzianowski2019}, we choose the GPT-2 LM as our backbone and use the LM to model both the belief state and the response.

\subsection{Model Representation}
\label{sec:model-repres}

The training instances for an LM-based task-oriented dialogue system can be considered as tuples $(c, b, d, r)$, where $c$ is the context (i.e., a concatenation of all previous utterances in the dialogue – both system's and user's), $b$ is the system's belief state (used to query the database), $d$ are the database results, and $r$ is the system's response. 

In our case, the dialogue system handles multiple domains and the belief state is a set of pairs (\emph{domain name}, \emph{domain belief}), where the \emph{domain belief} is an assignment of values into slots, i.e., a set of pairs (\textit{slot name}, \textit{value}) (see \exampleref{ex:augpt_format}). Similarly, the database results $d$ are a set of pairs (\textit{domain name}, \textit{domain database results}), where the \textit{domain database results} are an ordered list of entities returned by the database. We further define the \emph{database result counts} $d_c$ denoting the number of results in $d$ for each domain.

Ideally, we would like our system to model the probability distribution over possible responses conditioned on the context $p(r|c)$. To simplify computation and model external database queries, we factorize this distribution as follows:
\begin{equation}
\begin{split}
    p(r|c) &= \sum_{d} p(r|d,c) p(d|c) \\
           &= \sum_{d} \sum_{b} p(r|d,b,c) p(d|b) p(b|c) \\
           &= \sum_{b} p(r|\textit{Query}(b),b,c) p(b|c)\,,
\end{split}
\end{equation}
where $p(d|b)$ is a deterministic distribution over the database results, and \textit{Query} is a function returning database results.

Using this factorization allows the model to process the context, query the database and generate a response based on database results.
However, generating responses directly would result in data sparsity issues with rare tokens (e.g., venue names or reference numbers).
To maximally reuse the training samples, we choose to train our model on \emph{delexicalized responses}  denoted $\bar{r}$, where slot values are replaced with placeholders \cite{wen2015}. During inference, the responses are lexicalized back deterministically using the belief state and the database results. We assume perfect lexicalization, i.e., always being able to lexicalize the response $\bar{r}$ back based on $d$ and $b$.\footnote{We found in our experiments on the MultiWOZ data (see Section~\ref{sec:experiments}) that this assumption was almost always fulfilled.}

Both the database lookup and the lexicalization are deterministic, and the delexicalized response $\bar{r}$ does not depend on the database results $d$, but only on their counts $d_c$. Therefore, the distribution $p(r|d,b,c)$ is equal to the distribution $p(\bar{r}|d_c,b,c)$, and by maximizing its likelihood we are achieving the goal of maximizing the likelihood of $p(r|c)$.

We use the same language model $\hat{p}$ to model the belief state and to generate the delexicalized prediction. That is,
\begin{eqnarray}
    p(\bar{r}|d_c,b,c) & \approx & \hat{p}(\bar{r}|d_c,b,c,\theta) \\
    p(b|c) &  \approx&  \hat{p}(b|\emptyset,\emptyset,c,\theta)\,,
\end{eqnarray}
where we denote the model's parameters as $\theta$.

In the MultiWOZ dataset \cite[see Section~\ref{sec:experiments}]{budzianowski2018,eric2019}, responses are delexicalized by replacing concrete values with placeholder tokens of the form \textit{domain\_slot}. For better generalization across domains, we chose to only use \textit{slot} instead
as responses rarely involve more than one domain.
We train our model to predict the \textit{active domain} by outputting it first in the belief state (remaining domains follow in lexicographical order).
The predicted active domain is then used during lexicalization.\footnote{A disadvantage of this approach is that we cannot determine the active domain if the belief state is empty. However, in such a case the lexicalization would fail anyway, so the system's performance is not affected by this decision.}

\begin{example}
\begin{mdframed}[style=ExampleFrame]
Belief state: train \{ leave at=15:30, \\ 
\-\hspace{10pt}arrive by=17:15 \}, \\
\-\hspace{10pt}hotel \{ price range = cheap \} \\
DB: train 23 matches, hotel no match
\end{mdframed}
\caption{String format for \augpt's belief state and database result count\label{ex:augpt_format}.}
\end{example}

To fully exploit natural language pre-training of our LM, we represent the belief state and database result counts as strings containing as few special tokens as possible (see~\exampleref{ex:augpt_format}).

\subsection{Model Training}
\label{sec:model-traning}
Although parameters are shared for the belief state predictor and the delexicalized response predictor, the training objectives differ slightly. We use cross-entropy loss for both; response prediction uses  unlikelihood loss \cite{welleck2019,li_dont_2020} as an additional objective. Unlikelihood loss penalizes repeated tokens, which helps the model avoid repetitions and increases output diversity.

To help the model learn a better internal representation from the data, we employ additional auxiliary tasks. Similarly to \citet{devlin2019} and \citet{peng2020}, we train a binary classifier to detect dialogue inconsistencies. In each training batch, we corrupt half of the samples by randomly applying one or more of the following changes with the same probability:
\begin{compactenum}
    \item We replace the belief state $b$ with another belief state, sampled uniformly randomly from the training data.
    \item We replace the delexicalized response $\bar{r}$ with a different randomly chosen one. If this change is applied in combination with the first one, the delexicalized response and the belief state are taken from the same random sample.
    \item A different valid value is uniformly sampled for each slot in the belief state. In this case, the domain names and domain order are unchanged (i.e., the active domain is the same).
\end{compactenum}
\vspace{\parsep}

The first two changes are identical to \citet{peng2020}. The third one is a new one which we find very useful 
-- it is much more challenging to detect if the belief state was changed when the domain stays the same.
Consistency detection employs an affine binary classifier
on top of last response token logits, trained using binary cross-entropy (BCE).

We also experiment with additional two classifiers predicting the user intent and the system action. These are implemented as two fully-connected layers attached to the last context token and the last database result token logits, respectively. 
However, based on our experimental results (see Table~\ref{tab:ablation_comparison}), we decided not to use these tasks in the final model.

We train the whole pipeline by optimizing the non-weighted sum of individual component losses, i.e., cross-entropy for belief state and response prediction, unlikelihood loss for the response, and BCE for consistency detection.

\subsection{Response Generation}
For each user input, the system goes through several stages (see Figure~\ref{fig:pipeline}):
(1) Previous dialogue context is passed to the LM, which greedily generates the string representation of the belief state. (2) The belief state is parsed and passed to the database handler. (3) The database handler  returns a set of results for each domain. (4) A string representation of database result counts is created 
(see \exampleref{ex:augpt_format}). (5) The context, belief state and database results are concatenated and passed again to the LM. We use nucleus sampling \cite{holtzman2019} to generate the delexicalized response.\footnote{We found nucleus sampling useful for generating the response since it increases diversity, but we prefer greedy decoding for the belief state with a fixed structure.} (6) Placeholders in the delexicalized response are replaced by values from the database results and the belief state. 

\subsection{Data Augmentation}
Following its successful usage in other NLP tasks, \cite{konstas_neural_2017,elder_shape_2020}, we experiment with data augmentation using paraphrases.
In our setup, we generate multiple paraphrases for each training utterance and use them to augment the training data.
This way, we effectively increase the variability of the data.

Various data-driven approaches for paraphrasing were proposed, the majority of them corpora-based \cite{madnani2010generating}.
Recently, machine translation systems showed strong performance in generating paraphrases using back-translation \cite{sennrich2016,edunov2018,federmann2019multilingual}, i.e., translating an English text into an intermediate language and then translating the result back into English.
We use two different Transformer-based machine translation systems to paraphrase our data.
We used \citet{edunov2018}'s system with French and the system of \citet{machavcek2020elitr,DBLP:journals/corr/abs-2104-05688} with additional 40 pivot languages.
Based on empirical analysis of translation quality, we chose 10 pivot languages for our data -- we obtain 10 different paraphrases for each input utterance.\footnote{Pivot languages used:  Albanian, Arabic, Bulgarian, Bosnian, French, German, Russian, Spanish, Slovak, Swedish.}
When training, we choose the input user utterance uniformly at random from the set of all 10+1 variants of the utterance (backtranslation outputs and the original one).

\section{Experiments}
\label{sec:experiments}
\begin{table*}[htbp]
    \centering \small
    \begin{tabular}{l|ccc|ccc}
      \toprule
      & \multicolumn{3}{c|}{MultiWOZ 2.0} & \multicolumn{3}{c}{MultiWOZ 2.1} \\
      method & inform & success & BLEU & inform & success & BLEU \\
      \midrule
      Human & 91.0 & 82.7 & -- & 86.3 & 79.1 & -- \\
      \midrule
      \textbf{\Augpt} & 83.1 & 70.1 & 17.2 & 83.5 & 67.3 & 17.2 \\
      SOLOIST \cite{peng2020} & 85.5 & 72.9 & 16.5 & -- & -- & -- \\
      SimpleTOD \cite{hosseini2020} & 84.4 & 70.1 & 15.1 & 85.0 & 70.5 & 15.2 \\
      LABES-S2S \cite{zhang2020end2end} & -- & -- & -- & 78.1 & 67.1 & 18.3 \\
      DAMD \cite{zhang2019} & 76.3 & 60.4 & 16.6 & -- & -- & -- \\
      MD-Sequicity \cite{zhang2019} & 86.6 & 71.6 & 16.8 & -- & -- & -- \\
      LAVA \cite{lubis-etal-2020-lava} & 91.8 & 81.8 & 12.0 & -- & -- & -- \\
      \bottomrule
  \end{tabular}
  \caption{Comparison with previous works on the MultiWOZ dataset (see Section~\ref{sec:corpus-based} for a description of the metrics). \emph{MD-Sequicity} is a variant of \citet{lei2018}'s model, extended for a multi-domain setting.}
  \label{tab:multiwoz_sota_comparison}
\end{table*}
\begin{table*}[htbp]
    \centering \small
    \begin{tabular}{l|ccc|ccc|cc}
      \toprule
       &  & & & \multicolumn{3}{c|}{inform} & \multicolumn{2}{c}{turn} \\
      method & complete & success & book & P & R & F1 & succ & all \\
      \midrule
      \textbf{\Augpt} & 89.4 & 60.1 & 85.7 & 64.5 & 82.1 & 70.3 & 12.7 & 14.6 \\
      DAMD \cite{zhang2019} & 39.5 & 34.3 & 51.4 & 60.4 & 59.8 & 56.3 & 15.8 & 29.8\\
      Sequicity \cite{lei2018} & 23.1 & \phantom{0}9.8 & \phantom{0}4.1 & 33.0 & 32.7 & 29.9 & 12.2 & 32.6 \\
      \bottomrule
  \end{tabular}
  \caption{ConvLab evaluation comparison with other works (see Section~\ref{sec:convlab-eval} for a description of the metrics).}
  \label{tab:multiwoz_convlab_comparison}
\end{table*}

\subsection{Datasets}
As our primary dataset, we use \multiwozn, a de-noised version of \multiwoz \cite{budzianowski2018}.
We also used the 2.0 version
to compare to previous works.
The dataset contains 7 distinct domains (all related to tourist information, e.g., hotels, restaurants) and 10,438 dialogues, 7,032 of which are multi-domain.

We experiment with pre-training our model on additional datasets.
For the pre-training phase, we use \taskmaster \cite{byrne2019} and \schema \cite{rastogi2019}.\footnote{There are also other large-sized task-oriented datasets such as MetalWOZ \cite{lee2019multi}, however, 
their annotation is not detailed enough for our setup.}
Both \taskmaster and \schema are multi-domain, task-oriented, large dialogue corpora consisting of 12,215 and 22,825 dialogues, respectively.
\taskmaster was obtained using the Wizard-of-Oz and self-dialogue methods, while the collection of \schema is somewhat artificial -- humans are only employed to paraphrase machine-generated utterances.

\subsection{Data Preprocessing}\label{sec:preprocessing}

Although the MultiWOZ 2.1 dataset was collected by humans, it contains a lot of inconsistencies. We hypothesize that when using only \textit{clean} samples which are consistent with the database, the benefit of using higher quality training data outweighs the decrease in the number of training samples. This claim is further supported by experiments (see Section~\ref{sec:ablation}). To filter the training data, we choose only those dialogues where the annotated dialogue goal corresponds with the turn-level annotated data. When using the \emph{clean} samples, we omit about 30\% of the training data.

To effectively combine all our datasets, we unified the data ontologies.
Since the datasets use different naming conventions (e.g., \code{leaveAt} vs.\ \code{leave\_at}) and different domain and slot names to 
describe the same concepts (e.g., \code{restaurant-food} vs.\ \code{restaurant-type}),
we manually designed a mapping between domain and slot names. Notably, we decided to rename some slots so they use natural language tokens, as we base our model on the GPT-2 LM which is pre-trained on natural language texts (e.g. ``\code{leaveAt}'' $\rightarrow$ ``\code{leave at}'').
Our final ontology that unifies all three datasets contains 22 domains and 135 slots.

We use our own implementation of delexicalization, which directly produces our belief state string representation (see Section~\ref{sec:model-repres} and \exampleref{ex:augpt_format}).

\subsection{Training Details}

We implement our model in PyTorch \cite{pytorch}, based on GPT-2-\emph{small}.
It uses 12 layers with a size of 768.
For all auxiliary tasks, we use a dropout of 0.1 with label smoothing 0.1.
We use the AdamW optimizer \cite{loshchilov2017decoupled}.
The finetuning runs for 8 epochs on the \multiwozn data when all the training examples are used, and for the same number of minibatches when using only \emph{clean} samples.
The training takes less than one day when using 4 GPUs.

\begin{table*}[t]
    \centering\small
    \begin{tabular}{lcccccc}
      \toprule
      & Average & Success & Success  & NLU &  Response & \\
      Method & Success & w/ DB & w/o DB & score & appropriateness & Turns \\
      \midrule
      Baseline & 69.6 & 56.8 & 82.4 & 4.34 & 4.18 & 18.5 \\
      Winner & \textbf{74.8} & \textbf{70.2} & 79.4 & \textbf{4.54 }& \textbf{4.47} & 18.5 \\
      Our submission & 72.3 & 62.0 & \textbf{82.6} & 4.53 & 4.41 & \textbf{17.1} \\
      \bottomrule
    \end{tabular}
    \caption{Human evaluation results obtained during the DSTC9 shared task using Amazon Mechanical Turk. Note that only 4 out of 10 submissions outperformed the Baseline according to the average success metric.}
    \label{tab:human}
\end{table*}

\subsection{Corpus-based Evaluation}
\label{sec:corpus-based}

To compare with previous results on MultiWOZ, we evaluate the model performance with a set of corpus-based intrinsic metrics on both versions of the data.
For MultiWOZ 2.0, we use the original delexicalization used by compared baselines \cite{peng2020,hosseini2020,zhang2019}. For MultiWOZ 2.1, we use our own delexicalization.
We employ the original evaluation scheme by \citet{budzianowski2018}, which provides two metrics -- the \emph{inform rate} and the \emph{success rate}. 
The \emph{inform rate} is the percentage of dialogues in which the system mentioned a name or ID of an entity which does not contradict the current dialogue state and the user's goal, whereas the \emph{success rate} is the percentage of dialogues in which the system outputted all the requested information. 
Moreover, we compute BLEU \cite{papineni2002} between the generated system utterances and the ground truth to get an approximation of the output fluency.

\subsection{ConvLab~2 Evaluation}
\label{sec:convlab-eval}

We use the ConvLab~2 platform \cite{zhu2020} for automatic evaluation with a simulated user agent.
We run the evaluation component 1,000 times, i.e.\ on 1,000 simulated conversations.
The agent mimics user behavior, interacts with the system under evaluation, and computes multiple metrics: 
The \emph{complete rate} reflects the ratio of dialogues that are completed, i.e.\ all the user requests have been met.
The \emph{success rate} computes the percentage of dialogues which are successful, meaning the system captures correct informed entities and provides a valid booking if requested.
Finally, the \emph{book rate} is the proportion of dialogues where the system was able to book the correct entity (hotel, restaurant, train) if it was asked to.
We also compute \emph{precision, recall} and \emph{F1 score} for the informed entities and the average number of turns in the dialogue.

\subsection{Human Evaluation and Error Analysis}
\label{sec:human-eval}

Thanks to our participation in the DSTC9 task-oriented dialogue shared task \cite{gunasekara2020overview,li_multi-domain_2021}, a variant of our model (without pre-training on additional dialogue datasets, see Table~\ref{tab:ablation_comparison}) was selected for evaluation by human judges on the Amazon Mechanical Turk platform.\footnote{The selection was done based on ConvLab~2 performance, but probably used a different version of the tool and thus arrived at different results -- the chosen variant is not the best one according to our own measurements. \label{fn:dstc-eval}} The judges communicated with the agent in natural language and rated the system afterward with respect to the success/failure of the dialogue, language understanding score, and response appropriateness. Information provided by the system was additionally checked for consistency with the database, and the average of success rates given by the judges and by database grounding is used as the main metric.

In addition to the crowdsourced evaluation, we perform a detailed in-house error analysis based on human interactions with our final system. Expert annotators followed randomly chosen dialogue goals accompanying the MultiWOZ test set and recorded any incorrect system behavior.

\section{Results}
\label{sec:results}

We first discuss quantitative results for both corpus-based and crowdsourced human evaluation, then
include a qualitative analysis of the model behavior.

\subsection{Corpus-based Evaluation on MultiWOZ}

Table \ref{tab:multiwoz_sota_comparison} shows a comparison between our methods and current state-of-the-art systems (cf.~Section~\ref{sec:related}).
Since some of the compared methods do not provide results with on \multiwozn, we report results on both \multiwoz and MultiWOZ 2.1.
As we can see, \augpt compares favorably to other approaches.
The chosen variant of our model is not the best-scoring variant on corpus-based metrics (see Table~\ref{tab:ablation_comparison}). It was chosen based on the ConvLab evaluation, which may not be optimal for corpus-based evaluation.
LABES-S2S produces higher BLEU scores, which would indicate a better fluency of the model, but scores lower on inform and success rates.
LAVA, SOLOIST, SimpleTOD, and MD-Sequicity, on the other hand, provide slightly higher inform and success scores while doing worse in terms of fluency.

Table~\ref{tab:multiwoz_convlab_comparison} shows a comparison with two other models in the ConvLab evaluation scheme with a simulated user. The compared systems were chosen because they both implement fully trainable end-to-end methods. Our system outperforms both compared systems by a wide margin. Our model is able to perform well not just in a single-turn response generation scenario, but over the course of the whole dialogue. As the example of DAMD shows, this is not always guaranteed.

\begin{table*}[tp]
    \centering\small
    \begin{tabular}{l|ccc|ccc|ccc|cc}
      \toprule
        & \multicolumn{3}{c|}{MultiWOZ 2.1} & \multicolumn{8}{c}{ConvLab 2}  \\
       & \multicolumn{3}{c|}{} & \multicolumn{3}{c}{} & \multicolumn{3}{c}{inform} & \\
      method \text& inf & suc & BLEU & \hspace{-1mm}comp\hspace{-1mm} & \hspace{-1mm}suc\hspace{-1mm} & book & P & R & F1 & turns \\
      \midrule
      \textbf{\augpt} & 83.5 & 67.3 & 17.2 & \textbf{89.4} & \textbf{60.1} & 85.7 & 64.5 & \textbf{82.1} & 70.3 & 14.6 \\
    \midrule
    w/o. unlikelihood & 84.1 & 66.9 & 17.1 & 89.2 & 59.3 & \textbf{90.8} & 63.9 & 81.6 & 69.5 & 14.6 \\
    w/o. clean & 81.9 & 64.0 & 15.8 & 85.0 & 57.7 & 85.6 & 65.6  & 79.1 & 69.6 & 14.5 \\
    w/o. unlikelihood, w/o. clean & \textbf{86.5} & \textbf{69.1} & \textbf{17.5} & 85.9 & 58.4 & 81.3 & 62.2 & 79.8 & 67.5 & \textbf{14.1} \\
    w. all auxiliary & 83.1 & 66.2 & 17.0 & 88.7 & 59.2 & 86.0 & 64.6 & 81.1 & 69.9 & 14.4 \\
    \midrule
    w/o. pre-training & 81.0 & 62.7 & 15.1 & 88.1 & 59.8 & 83.7 & \textbf{68.1} & 80.9 & 72.1 & 15.6 \\
    w/o. back-translations & 79.8 & 61.7 & 15.2 & 88.9 & 58.2 & 87.4 & 68.0 & 81.6 & \textbf{72.2} & 14.9 \\
    w. old consistency & 81.4 & 65.8 & 17.0 & 85.5 & 57.8 & 86.0 & 65.2 & 80.0 & 69.8 & 14.6  \\
    w/o. consistency & 81.9 & 64.5 & 16.3 &  86.4 & 57.1 & 84.1 & 66.3 & 81.2 & 70.9 & 14.6 \\
    \bottomrule
  \end{tabular}
  \caption{Ablation study (inf = inform, suc = success, book = book rate; see Section~\ref{sec:corpus-based}). The model version with the best ConvLab~2 success rate is chosen as our best model. Variants are denoted with their respective modifications compared to the default: ``w/o.\ unlikelihood'' = unlikelihood loss was not used; ``w/o.\ clean'' uses all training samples as opposed to using only the ones consistent with the database; ``w/o.\ pre-training'' = Taskmaster-1 and Schema-Guided datasets were not used for training (this variant was selected for human evaluation); ``all auxiliary'' = using two additional auxiliary tasks (see the Method section); ``w/o.\ consistency'' = dialogue consistency task is not used; ``old consistency'' refers to the consistency task by \citet{peng2020} (see the Section~\ref{sec:model-traning}).}
  \label{tab:ablation_comparison}
\end{table*}

\subsection{Human Evaluation}

We show the DSTC9 human evaluation results in Table~\ref{tab:human}.
Our model outperformed the baseline on the main average success metric (along with only 3 other submissions out of 10).
We did not match the winning submission on average success, but we achieved the best overall success rate based on human judgments (without checking consistency with the database).
This suggests that the main drawback of our model lies in database grounding.
This could have been caused by our new delexicalization (Section~\ref{sec:preprocessing}).
Finally, our model needed the smallest number of turns on average to complete the dialogues.

Note that the model variant chosen for human evaluation did not include pre-training on additional data (cf.~Section~\ref{sec:human-eval} and Footnote~\ref{fn:dstc-eval}), which could improve performance in human evaluation.

\subsection{Error Analysis}\label{sec:quality-analysis}

\begin{example}[t]
\small
\setlength{\tabcolsep}{2pt}
\begin{mdframed}[style=ExampleFrame]
\begin{tabularx}{\textwidth}{rX}
[...] \\
\bf user: & \it I want to make a booking at sitar tandoori \\
\bf belief: & restaurant \{name = sitar tandoori\} \\
& train \{arrive by = 18:00, day = sunday, departure = cambridge, destination = london liverpool street, people = 1\} \\
\bf system: & What day would you like the booking for? \\
\bf user: & \it the same day, 13:30 \\
\bf belief: & restaurant \{day = sunday, name = sitar tandoori, people = 1, time = 13:30\} \\
& train \{arrive by = 18:00, day = sunday, departure = cambridge, destination = london liverpool street, people = 1\} \\
\bf system: & Booking was successful. The table will be reserved for 15 minutes. Reference number is: 00000200. \\
\end{tabularx}
\end{mdframed}
\caption{A dialogue where the system did not ask for necessary information. The user previously booked a train trip for one person (cf.~the belief state in the first turn shown). When they went on to book a restaurant, the system took over the number of persons from the train ride and did not ask again.\label{ex:error1}}
\end{example}

\begin{example}[ht!]
\small
\setlength{\tabcolsep}{2pt}
\begin{mdframed}[style=ExampleFrame]
\begin{tabularx}{\textwidth}{rX}
\bf user : & I am looking for a cheap \textit{chinse} restaurant. \\
\bf belief: & restaurant \{food = chinese, price range = cheap\} \\
\bf system: & I have 4 cheap chinese restaurants. What area of town would you like to dine in?\\
\bf user: & downtown   \\ 
\bf belief: & restaurant \{area = centre, food = chinese, price range = cheap\} \\
\bf system: & I have 3 cheap chinese restaurants in the centre. Would you like to book at Charlie Chan?\\
 & ... \\

\end{tabularx}
\end{mdframed}
\caption{A snippet from a challenging but successful dialogue. The system deals with input noise (“chinse”) as well as with paraphrases (“downtown”). \label{ex:success}}
\end{example}
Our expert annotators evaluated 130 dialogues in total, 50 of which contained at least one error.
However, in most cases, the system was able to recover from the errors, resulting in an overall success rate of 86.9\% (i.e., 17 unsuccessful dialogues).
The purpose of this analysis was to identify different types of errors occurring during full dialogues.

By far the most common error  (21 counts) were \emph{hallucinated values}, i.e., lack of grounding for some of the information provided (see the end of \exampleref{ex:error1}).
Another frequent error type is \emph{missing information} (5 counts), i.e., not asking for information that is required (and e.g. reusing information from a different domain without user confirmation).
Example~\ref{ex:error1} also demonstrates another common error type, which is \emph{bad domain} (4 counts). Here, the system does not react to the user's request for a different search (hotels instead of attractions in the example). This might be caused by a less frequent wording of the request in the given context, and usually gets resolved by rephrasing the request.

The analysis also found many examples of correct system behavior in non-trivial cases.
As illustrated in Example~\ref{ex:success},
the model is able to deal with paraphrases and is robust to a certain amount of noise in the data. Specifically, it handles typos, various time formats, etc.
Interaction between domains is also successful in most cases -- the system is able to resolve references to another domain's belief state (e.g., make a booking for the same group of people as done before in a different venue).

\section{Ablation Study}
\label{sec:ablation}

We tested many variants of our method with different combinations of components to evaluate their contributions. The results are presented in Table \ref{tab:ablation_comparison}.
Namely, we are interested in the following components:
\textbf{(1)} unlikelihood loss, \textbf{(2)} auxiliary tasks, \textbf{(3)} data augmentation, \textbf{(4)}  modified consistency task and \textbf{(5)} unclean data filtering.

We can see that all proposed contributions which are a part of our final system, except for the unlikelihood training, have a positive effect on the system performance.
In the ConvLab evaluation, our final system performs best. 
Removing either pre-training or back-translations decreases BLEU, inform and success rates substantially. Furthermore, we notice the positive effect of using our improved consistency detection task over the one used in SOLOIST \cite{peng2020}, which in turn scores better than no consistency detection. 

Training on all data as opposed to using only “clean” samples clearly reduces performance. On the other hand, unlikelihood training improves performance only in ConvLab while causing a performance drop in corpus-based metrics. This can be caused by the fact that the unlikelihood training promotes diversity and reduces repetitions on the token level, and thus does not play well with corpus-based evaluation. 
We did not notice any increase in performance when the user intent prediction and system action prediction auxiliary tasks were used (cf.\ Section~\ref{sec:model-traning}).
The reason for this behavior could be that the model learns to represent the actions well enough implicitly, without the need for these additional objectives.
Therefore, these tasks are not a part of our final model.

\section{Conclusions \& Future Work}
We present a dialogue modeling pipeline based on the pre-trained GPT-2 language model.
\Augpt uses modified training objectives and employs data augmentation to increase the diversity of generated utterances.
Our experiments show that the proposed approach outperforms baselines and is competitive with state of the art on the MultiWOZ dataset.
We also run a series of ablation experiments to assess the individual contributions of the modifications.
According to our detailed ablation study, 
training data augmentation using back-translation via multiple languages and a modified auxiliary training objective for dialogue consistency detection are the features that contribute most to our system's performance.
Additionally, we perform a qualitative analysis of the outputs to give a better insight into our model behavior.

In the future, we plan to construct a latent representation of the belief state and optimize it jointly with the language model. We will replace the deterministic lexicalization with a trainable alternative, and possibly even integrate the database module into the model. To improve the transfer to new domains, we will learn a domain embedding and optimize it jointly with the model, unifying all datasets.

\subsection*{Acknowledgments}
This work was supported by the Charles University GAUK grant No.~302120 and No.~373921, the SVV project No.~260575, and the Charles University project PRIMUS/19/SCI/10.
Jonáš Kulhánek was supported by the European Regional Development Fund under the project Robotics for Industry 4.0 (reg.~no.\ CZ.02.1.01/0.0/0.0/15\_003/0000470). Additional computational resources were supplied by the project ``e-Infrastruktura CZ'' (e-INFRA LM2018140) provided within the program Projects of Large Research, Development and Innovations Infrastructures.

\bibliography{bibliography}
\bibliographystyle{acl_natbib}

\clearpage
\appendix
\section{Additional Results}
\label{sec:appendix}
\subsection{Detailed Error Analysis}
Our expert annotators evaluated 130 dialogues in total, 50 of which contained at least one error.
However, in most cases, the system was able to recover from the errors, resulting in an overall success rate of 86.9\% (i.e., 17 unsuccessful dialogues).
The purpose of this analysis was to identify different types of errors occurring during full dialogues.
The annotators were familiar with the model architecture and were instructed to categorize the errors according to the cause of the problem.
Specifically, they identified which component caused the respective error
and annotators categorized the errors into more specific types.

The overall results are given in Table~\ref{tab:interact_eval}.
We observe that the most common reason for a failed dialogue is an error related to the belief state (30 errors, 10 failed dialogues).
Also, although policy errors happen relatively often (21x), they rarely cause the whole dialogue to fail (2 dialogues).
We observe that we have a slightly higher number of successful dialogues compared to the 82.6\% success rate (without checking database consistency) found in human evaluation (cf.\ Table~\ref{tab:human}).
The most likely cause is that our expert annotators were more motivated to recover from erroneous system behavior and finish the dialogue.

Fine-grained error types identified by annotators are given in Table~\ref{tab:detailed_eval} and Examples~\ref{ex:error1},~\ref{ex:error2} and~\ref{ex:success}.
To extend the analysis from Section~\ref{sec:quality-analysis}, we include another frequent error type -- \emph{missing information} (5 counts), i.e., not asking for information that is required (\exampleref{ex:error2}). In this case, the system uses information from a different domain without the user explicitly confirming this.
A most probable cause of this is that most instances of the training data carry over the information.

\begin{table*}[t]
    \centering\small
    \begin{tabular}{lc|ccccc}
      \toprule
      & Erroneous & BS & DB & Policy & Other \\
      & Dialogues  & Error & Error & Error & Error \\
      \midrule
      All dialogues & 50  & 30 & 4 & 21 & 6 \\
      Unsuccessful dialogues & 17 & 10 & 3 & 2 & 2 \\ 

      \bottomrule
    \end{tabular}
    \caption{Interactive  analysis performed by human evaluators using 130 prepared dialogue goals. 17 of these dialogues contained an error that caused the dialogue to fail. We show summary statistics regarding the number of respective error sources (BS = belief state, DB = database).
    Note that some of the dialogues contain more than one error.}
    \label{tab:interact_eval}
\end{table*}

\begin{example}[tpbh]
\small
\setlength{\tabcolsep}{2pt}
\begin{mdframed}[style=ExampleFrame]
\begin{tabularx}{\textwidth}{rX}
\bf user: & \it I'm looking for a theatre in Cambridge \\
belief: & attraction \{type = theatre\} \\
system: & There are 5 theatres in Cambridge. Is there an area of town you prefer? \\
\bf user: & \it no I don't care \\
belief: & attraction \{type = theatre\} \\
system: & I recommend Adc Theatre in the centre. Would you like more information? \\
\bf user: & \it no, I'm also looking for some place to stay \\
belief: & attraction \{type = theatre\} \\
\bf system: & I have 5 theatres in the centre. Do you have a price range? \\
\bf user: &  \it I'm looking for a place to stay \\
belief: & attraction \{type = theatre\} \\
\bf system: & There are 5 theatres in the centre. Do you have a preference for area? \\
\bf user: & \it I'm looking for a hotel \\
belief: & hotel \{type = hotel\} \\ 
& attraction \{type = theatre\} \\
\bf system: & \it There are 9 hotels in the north. Do you have a price range?\\
\end{tabularx}
\end{mdframed}
\caption{Dialogue sample with a bad domain focus and a hallucination. First, the system ignores that the user switched from searching for a theater to searching for a hotel. After accepting the new domain, the system replies with hotels “in the north” even though the user did not specify.\label{ex:error2}}
\end{example}

\begin{table*}[t]
\centering\small
\begin{tabularx}{\textwidth}{lrcX}
\toprule
Type & Count & Source & Description \\
\midrule
Hallucinated values & 21 & BS/Policy & Used a slot value in the reply that is not grounded in the DB nor in the context \\
Wrong lexicalization & 6 & Policy & Repeats the same value in a list of choices during lexicalization  \\
Missing information & 5 & Policy & Makes booking while not all information is specified \\
Ignored input  & 5& BS & Keeps asking for information that was provided \\
Bad domain     & 4 & BS  & Fails to focus on the correct domain \\
False response & 4 & Policy & States a different value of a slot than the value stored in DB \\
Repeated output& 3 & Policy & Repeats the same slot twice on the output \\
Failed booking & 3 & DB/Policy & Booking was unsuccessful due to DB mismatch\\
Other          &10 & BS/DB/P/Oth & (Various rare errors that could not be categorized) \\
\bottomrule
\end{tabularx}
\caption{Distribution of the most common error types encountered during the human evaluation of 130 dialogues. Absolute counts of errors in the 50 erroneous dialogues are shown. The total error count is 61 as some dialogues contained multiple errors. The most likely source of the error (cf.~Table~\ref{tab:interact_eval}) and a short description are given for each type.}
\label{tab:detailed_eval}
\end{table*}

\subsection{Individual Component Analysis}
We have conducted additional tests to obtain a deeper insight into each component's performance -- DST and NLG. We have evaluated the accuracy of the generated belief states. Joint accuracy, slot accuracy, and F1 score were used. Joint accuracy gives the percentage of successfully generated belief states -- with no error. Slot accuracy, on the other hand, is the average accuracy of correctly predicting the value for a domain-slot pair. To evaluate NLG, we compared the end-to-end system where the generated belief state is used to query the database and generate the response with a variant of the pipeline, where the ground-truth belief state and/or ground-truth database result counts were used. The BLEU \citep{papineni2002} and ROUGE-L \citep{lin2004rouge} scores were used for evaluation.

In Table~\ref{tab:component-analysis}, we can see the performance of each individual component of the system. One can notice that the performance of NLG is not decreased when we use the generated belief state instead of the oracle belief state. Since the belief state prediction is not perfect, this suggests that the model does not actually need belief states for generating the delexicalized response. However, when the real database result counts are used instead of oracle database result counts, the performance decreases, which implies that the database result counts are important for NLG.

\begin{table*}[htbp]
    \centering\small
    \begin{tabular}{l|cc|ccc|cc}
        & \multicolumn{2}{c|}{oracle} & \multicolumn{3}{c|}{DST} & \multicolumn{2}{c}{NLG} \\
        fine-tuned on & bs & db & joint acc. & slot acc. & F1 & BLEU & ROUGE-L \\
        \hline
        \multirow{3}{*}{MW 2.0} & \xmark & \xmark & \multirow{3}{*}{54.1} & \multirow{3}{*}{97.2} & \multirow{3}{*}{90.0} & 17.2 & 39.0 \\
        & \xmark & \cmark &  & & & 17.4 & 39.3 \\
        & \cmark & \cmark &  & & & 17.4 & 39.2 \\
        \hline
        \multirow{3}{*}{MW 2.1} & \xmark & \xmark & \multirow{3}{*}{56.5} & \multirow{3}{*}{97.2} & \multirow{3}{*}{90.6} & 17.4 & 38.6 \\
        & \xmark & \cmark &  & & & 17.6 & 38.8 \\
        & \cmark & \cmark &  & & & 17.6 & 38.8 \\
        \hline
    \end{tabular}
    \caption{Performance of DST and NLG components. Joint and slot accuracies, as well as slot values F1 score, are used to evaluate DST. For NLG, BLEU and ROUGE-L metrics are used. Apart from using the generated belief states and database counts, we also evaluate the components with oracle values. Note that models were pre-trained on Taskmaster-1 and Schema-Guided Dialogue datasets.}
    \label{tab:component-analysis}
\end{table*}
\end{document}